\documentclass[11pt]{article}
\usepackage{acl2016}
\usepackage{times}
\usepackage{bm}
\usepackage{graphicx,color,xcolor}
\usepackage{latexsym}
\usepackage{amsmath}
\usepackage{amsfonts}
\usepackage{dsfont}
\usepackage{multirow}
\usepackage{booktabs, caption}
\usepackage{tabularx}
\usepackage{listliketab}
\usepackage{lipsum} 
\usepackage{float}
\aclfinalcopy % Uncomment this line for the final submission
\def\aclpaperid{752} %  Enter the acl Paper ID here

\newcommand{\M}{\ensuremath{\mathbf{M}}}

\newcommand{\W}{\ensuremath{\mathbf{W}}}

\renewcommand{\c}{\ensuremath{\mathbf{c}}}

\newcommand{\e}{\ensuremath{\mathbf{e}}}

\newcommand{\h}{\ensuremath{\mathbf{h}}}

\newcommand{\sss}{\ensuremath{\mathbf{s}}}  % TIPA defines \s and LaTeX \ss!

\newcommand{\vv}{\ensuremath{\mathbf{v}}}
\newcommand{\calV}{\ensuremath{\mathcal{V}}}

\newcommand{\calX}{\ensuremath{\mathcal{X}}}

{%
\begin{list}{#1}{
\vspace{-\topsep}
\vspace{-\partopsep}
\setlength{\itemindent}{0cm}
\setlength{\rightmargin}{0cm}
\setlength{\listparindent}{0cm}
\settowidth{\labelwidth}{#1}
\setlength{\leftmargin}{\labelwidth}
\addtolength{\leftmargin}{\labelsep}
\setlength{\itemsep}{0cm}
}%
}%
{%
\end{list}
\vspace{-\topsep}
\vspace{-\partopsep}
}

{\begin{enumerate}%
}%
{\end{enumerate}}

\title{Incorporating Copying Mechanism in Sequence-to-Sequence Learning}
\def\fndaff{$^\ddag$}
\def \hku{$^\dagger$}
\author{Jiatao Gu\hku ~~~~ Zhengdong Lu\fndaff ~~~~ Hang Li\fndaff  ~~~~ Victor O.K. Li\hku
\\
{\hku {Department of Electrical and Electronic Engineering, The University of Hong Kong}} \\
{ \tt \{jiataogu, vli\}@eee.hku.hk}\\
{ \fndaff {Huawei Noah's Ark Lab, Hong Kong}}   \\
{ \tt \{lu.zhengdong, hangli.hl\}@huawei.com}\\
}

\date{}

\begin{document}

\maketitle

\begin{abstract}
We address an important problem in sequence-to-sequence (Seq2Seq) learning referred to as copying, in which certain segments in the input sequence are selectively replicated in the output sequence.  A similar phenomenon is observable in human language communication. For example, humans tend to repeat entity names or even long phrases in conversation. The challenge with regard to copying in Seq2Seq is that new machinery is needed to decide when to perform the operation. In this paper, we incorporate copying into neural network-based Seq2Seq learning and propose a new model called \textsc{CopyNet} with encoder-decoder structure.  \textsc{CopyNet} can nicely integrate the regular way of word generation in the decoder with the new copying mechanism which can choose sub-sequences in the input sequence and put them at proper places in the output sequence. Our empirical study on both synthetic data sets and real world data sets demonstrates the efficacy of \textsc{CopyNet}. For example, \textsc{CopyNet} can outperform regular RNN-based model with remarkable margins on text summarization tasks.   
   
\end{abstract}

\section{Introduction}
Recently, neural network-based sequence-to-sequence  learning (Seq2Seq)
has achieved remarkable success in various natural language processing (NLP) tasks, including but not limited to  Machine Translation~\cite{cho2014learning,bahdanau2014neural}, Syntactic Parsing~\cite{vinyals2015grammar}, Text Summarization~\cite{rush2015neural} and  Dialogue Systems~\cite{vinyals2015neural}. Seq2Seq is essentially an encoder-decoder model, in which the encoder first
transform the input sequence to a certain representation which can then transform the representation into the output sequence. 
Adding the attention mechanism~\cite{bahdanau2014neural} to Seq2Seq, first proposed  for automatic alignment in machine translation, has led to significant improvement on the performance of various tasks~\cite{shang2015neural,rush2015neural}. Different from the canonical encoder-decoder architecture, the attention-based Seq2Seq model revisits the input sequence in its raw form (array of word representations)  and dynamically fetches the relevant piece of information based mostly on the feedback from the generation of the output sequence.
%on content-based addressing. 
 
In this paper, we explore another mechanism important to the human language communication, called the ``copying mechanism". Basically, it refers to the mechanism that locates a certain segment of the input sentence and puts the segment into the output sequence. For example, in the following two dialogue turns we observe different patterns in which some subsequences (colored blue) in the response (\textsf{R}) are copied from the input utterance (\textsf{I}):

\vspace{2pt}
\noindent
\begin{tabularx}{\linewidth}{@{}>{\bfseries}l@{\hspace{.4em}}X@{}}
        \toprule
    \noindent {\sf \small I:}& \texttt{\small Hello Jack, my name is {\color{blue}Chandralekha}.}\\
    \noindent {\sf \small R:}& \texttt{\small Nice to meet you, {\color{blue}Chandralekha}.}            \\
    \midrule
    \noindent {\sf \small I:} &\texttt{\small This new guy {\color{blue} doesn't perform exactly as we expected}.}\\
    \noindent {\sf \small R:} &\texttt{\small What do you mean by "{\color{blue}doesn't perform exactly as we expected}"?}\\
    \bottomrule
\end{tabularx}
 \vspace{2pt}
 
Both the canonical encoder-decoder and its variants with attention mechanism rely heavily on the representation of ``meaning", which might not be sufficiently inaccurate in cases in which  the system needs to refer to sub-sequences of input like entity names or dates. In contrast, the copying mechanism is closer to the rote memorization in language processing of human being, deserving a different modeling strategy in neural network-based models. We argue that it will benefit many Seq2Seq tasks to have an elegant unified model that can accommodate both understanding and rote memorization. Towards this goal, we propose \textsc{CopyNet}, which is not only capable of the regular generation of words but also the operation of copying appropriate segments of the input sequence. Despite the seemingly ``hard" operation of copying, \textsc{CopyNet} can be trained in an end-to-end fashion. Our empirical study on both synthetic datasets and real world datasets demonstrates the efficacy of \textsc{CopyNet}.       
 
%\paragraph{RoadMap}~We will first  introduce the Seq2Seq learning framework briefly in Section 2. Then in Sections 3 and 4, we will discuss \textsc{CopyNet} and its learning. In Section 5 we report our empirical study of \textsc{CopyNet}, which is followed by related work (Section 6) and conclusion (Section 7).

%%%%%%%%%%%%%%%%%%%%%%%%%%%%%%%%%%%%%%%%%%%%%%%%%%%%
%%%%%%%%%%%%%%%%%%%%%%%%%%%%%%%%%%%%%%%%%%%%%%%%%%%%

%\newpage
\section{Background: Neural Models for Sequence-to-sequence Learning}
Seq2Seq Learning  can be expressed in a probabilistic view as maximizing the likelihood (or some other evaluation metrics~\cite{LiuYang}) of observing the output (target) sequence  given an input (source) sequence. % what's the meaning of r

% which is learning to generate the most proper sequence $Y$ that maximizes the conditional probability given a source sequence $X$, i.e. $\arg\max_{Y} P(Y|X)$. Recently, the quite a number of works are proposed to utilize the recurrent neural networks (RNNs) to directly learn to  generate the target sequence.

\subsection{RNN Encoder-Decoder}
RNN-based Encoder-Decoder is successfully applied to real world Seq2Seq tasks, first by \newcite{cho2014learning} and \newcite{sutskever2014sequence}, and then by~\cite{vinyals2015neural,vinyals2015pointer}. In the Encoder-Decoder framework, the  source sequence $X=[x_1, ..., x_{T_S}]$ is converted into a fixed length vector $\c$ by the encoder RNN, i.e.
\begin{equation} \label{eq:enc}
	\h_t = f(x_t, \h_{t-1});  \quad \c = \phi(\{\h_1, ..., \h_{T_S} \})
\end{equation} 
 where $\{\h_t\}$ are the RNN states, $\c$ is the so-called context vector, $f$ is  the dynamics function, and $\phi$ summarizes the hidden states, e.g.  choosing the last state $\h_{T_S}$.  In practice it is found that gated RNN alternatives such as LSTM~\cite{hochreiter1997long} or GRU~\cite{cho2014learning} often perform much better than  vanilla ones.
 
%{\color{red} With an RNN-based decoder, it naturally amounts to learning to 
%\emph{sequentially} generating the target sequence. 
%}
The decoder RNN is to unfold the context vector $\c$ into the target sequence, through the following dynamics and 
%observation model
prediction model:
 %receiving the previous symbol $y_{t-1}$, the previous hidden state $s_{t-1}$ and the context vector $c$, that is
   \vspace{-5pt}
 \begin{equation}\label{eq:dec}
 \begin{split}
 	 &\sss_t = f(y_{t-1}, \sss_{t-1}, \c)\\
	 &p(y_t|y_{<t}, X) = g(y_{t-1}, \sss_t, \c)
 \end{split}	
  \vspace{-5pt}
 \end{equation}
where $\sss_t$ is the RNN state at time $t$, $y_t$ is the predicted target symbol at $t$ (through function $g(\cdot)$) with $y_{<t}$ denoting the history $\{y_1, ..., y_{t-1}\}$. The prediction model is typically a classifier over the vocabulary with, say, 30,000 words.

\subsection{The Attention Mechanism}
The attention mechanism was first introduced to Seq2Seq~\cite{bahdanau2014neural} to release the burden of summarizing the entire source into a fixed-length vector as context. Instead, the attention
uses a dynamically changing context $\c_t$ in the decoding process. A natural option (or rather ``soft attention")  is to represent $\c_t$ as the weighted sum of the source hidden states, i.e.
\begin{equation} \label{eq:att}
	\c_t = \sum_{\tau=1}^{T_S}{\alpha_{t\tau} \h_{\tau}};\quad \alpha_{t\tau} = \frac{e^{\eta(\sss_{t-1}, \h_{\tau})}}{\sum_{\tau'} e^{\eta(s_{t-1}, \h_{\tau'})}}
\end{equation}
where $\eta$ is the function that shows the correspondence strength for attention, approximated usually with a multi-layer neural network (DNN). Note that in~\cite{bahdanau2014neural} the source sentence is encoded with a Bi-directional RNN, making each hidden state $\h_{\tau}$ aware of the contextual information from both ends.

% [maybe introduce the STM here]

%The attention mechanism allows the decoding with focus on different components of the source, and help the generation. However, the conventional mechanism only treats the source information as continuous real-valued vectors,  and influences the prediction in an indirect way.  It cannot directly generate the symbols of the source sequence, which sometimes is problematic when we only require accurate symbols in the source sentence rather than completely understand them to produce the reasonable sequence. Such symbols are usually out-of-vocabulary (OOV), for instance. 

\section{\textsc{CopyNet}}
From a cognitive perspective, the copying mechanism is related to rote memorization, requiring less understanding but ensuring high literal fidelity. From a modeling perspective, the copying operations are more rigid and symbolic, making it more difficult than soft attention mechanism to integrate into a fully differentiable neural model.
%the attention mechanism generally mimics the human's behavior of attentive reading the important part from the given conditions, and understanding it for some general objectives. 
In this section, we present \textsc{CopyNet}, a differentiable Seq2Seq model with ``copying mechanism", which can be trained in an end-to-end fashion with just gradient descent.

 \begin{figure*}[htbp]
   	\centering
          	\includegraphics[width=1\linewidth]{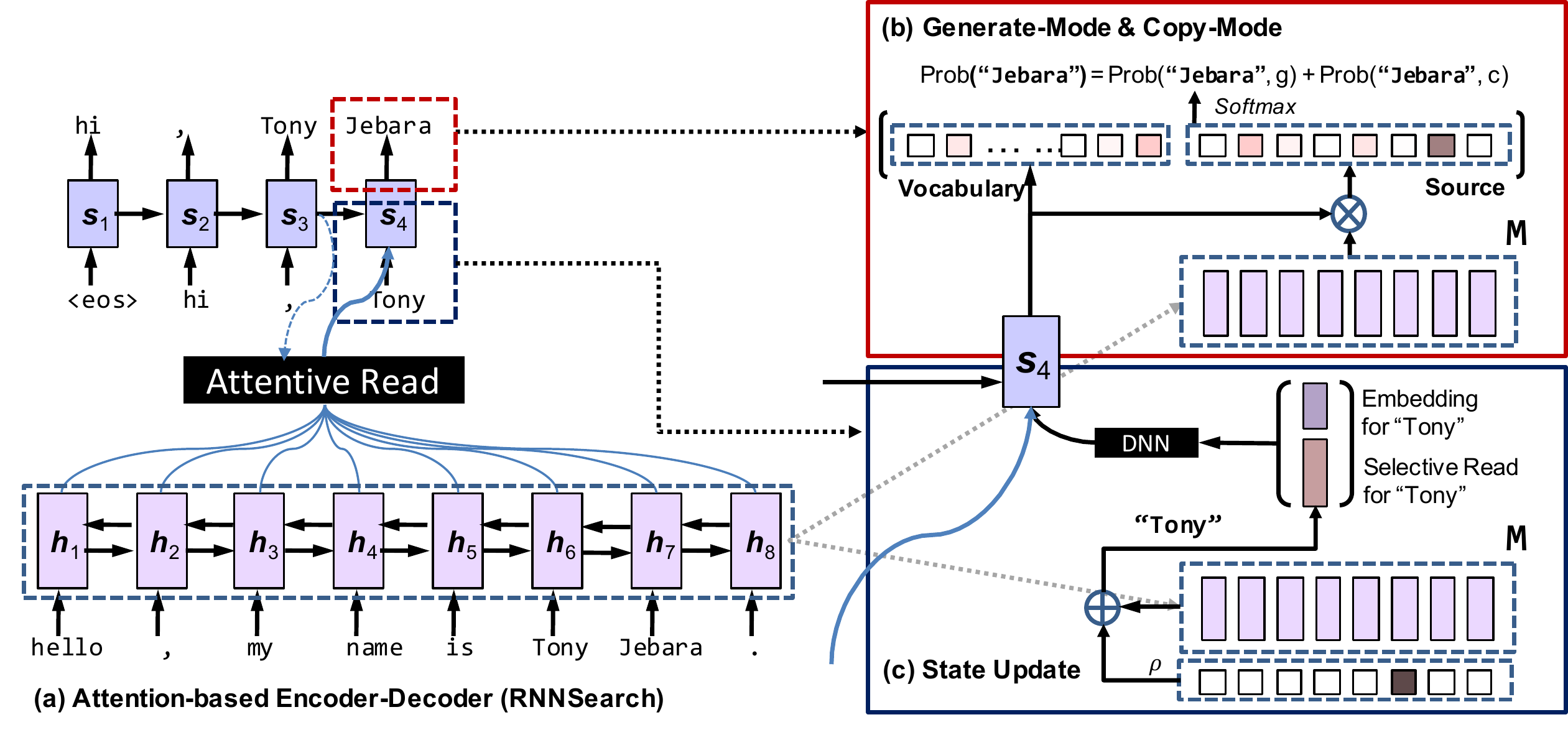} 
          	\caption{\label{model}The overall diagram of \textsc{CopyNet}. For simplicity, we omit some links for prediction (see Sections 3.2 for more details). }
          	\vspace{-7pt}
%          	\caption{\label{model}An example of \textsc{CopyNet} architecture based on the attention-based RNN encoder-decoder (a)  with the Copying Mechanism in the prediction module (b) and the state updating module (c). Note that the rightmost part is the added Copying Mechanism on the original model on the left. } 
  \end{figure*} 
  
\subsection{Model Overview}% across the source sequence, combining both content-based addressing and location-based addressing.
As illustrated in Figure~\ref{model}, \textsc{CopyNet} is still an encoder-decoder (in a slightly generalized sense). The source sequence is transformed by \textbf{Encoder} into 
% the source 
representation, which is then read by \textbf{Decoder} to generate the target sequence.
\vspace{-5pt}
\paragraph{Encoder:} Same as in~\cite{bahdanau2014neural}, a bi-directional RNN is used to transform the source sequence into a series of hidden states with equal length, with each hidden state $\h_t$ corresponding to word $x_t$. 
This new representation of the source, $\{\mathbf{h}_1, ..., \mathbf{h}_{T_S}\}$, is considered to be a short-term memory  (referred to as $\M$ in the remainder of the paper), which will later be accessed in multiple ways in generating the target sequence (decoding).

\paragraph{Decoder:}  An RNN that reads $\M$ 
% from the source STM, 
and predicts the target sequence. It is similar with the canonical RNN-decoder in~\cite{bahdanau2014neural}, with however the following important differences 
\begin{itemize}
	\vspace{-5pt}
    \item {\bf Prediction:}~ \textsc{CopyNet} predicts words based on a mixed probabilistic model of two modes, namely the \textbf{generate-mode} and the \textbf{copy-mode}, where the latter picks words from the source sequence (see Section \ref{sec:predict});
	\vspace{-5pt}
	\item {\bf  State Update:}~ the predicted word at time $t\hspace{-2pt}-\hspace{-2pt}1$ is used in updating the state at $t$, but \textsc{CopyNet} uses not only its word-embedding but also its corresponding location-specific hidden state in $\M$ (if any) (see Section \ref{sec:stateupdate} for more details);
	\vspace{-5pt}
	\item {\bf Reading $\M$:}~ in addition to the attentive read to $\M$, \textsc{CopyNet} also has``selective read" to $\M$, which leads to a powerful hybrid of content-based addressing and location-based addressing (see both Sections \ref{sec:stateupdate} and \ref{sec:reading} for more discussion).
	\vspace{-5pt}
\end{itemize}

%\begin{itemize}
%	\vspace{-5pt}
%    \item {\sf  Prediction:} \textsc{CopyNet} generates words through a mixture of two probabilistic models, namely the Generative-Mode and the Copy-Mode, where the latter picks a word from the source sequence (see  Section \ref{sec:predict} for more details);
%	\vspace{-3pt}
%	\item {\sf State Update:} the generated word at time $t$ will be used in updating the state or $t+1$, but \textsc{CopyNet} feeds not only the word-embedding but also the original location-specific vector in STM to the state updating function (see Section \ref{sec:stateupdate} for more details). 
%	\vspace{-3pt}
%	\item {\sf Reading STM:} in addition to the attentive read to STM, \textsc{CopyNet} has also ``selective read" to STM  built in its prediction and state update (see both Section \ref{sec:stateupdate} and \ref{sec:reading} for more discussion).
%\end{itemize}

\subsection{Prediction with Copying and Generation}
%\vspace{-2pt}
\label{sec:predict}
We assume a vocabulary $\calV=\{v_1, ..., v_N\}$, and use \textsc{unk} for any out-of-vocabulary (OOV) word. 
% Is it ok to have no introduction for OOV
In addition, we have another set of words $\calX$, for all the 
\emph{unique} words in source sequence $X=\{x_1, ..., x_{T_S}\}$.
Since $\calX$ may contain words not in $\calV$, copying sub-sequence in $X$ enables \textsc{CopyNet} to output some OOV words. In a nutshell, the instance-specific vocabulary for source $X$ is $\calV \cup \textsc{unk} \cup \calX$.

\begin{figure}[t!]
   	\centering
          	\vspace{-10pt}
          	\includegraphics[width=0.95\linewidth]{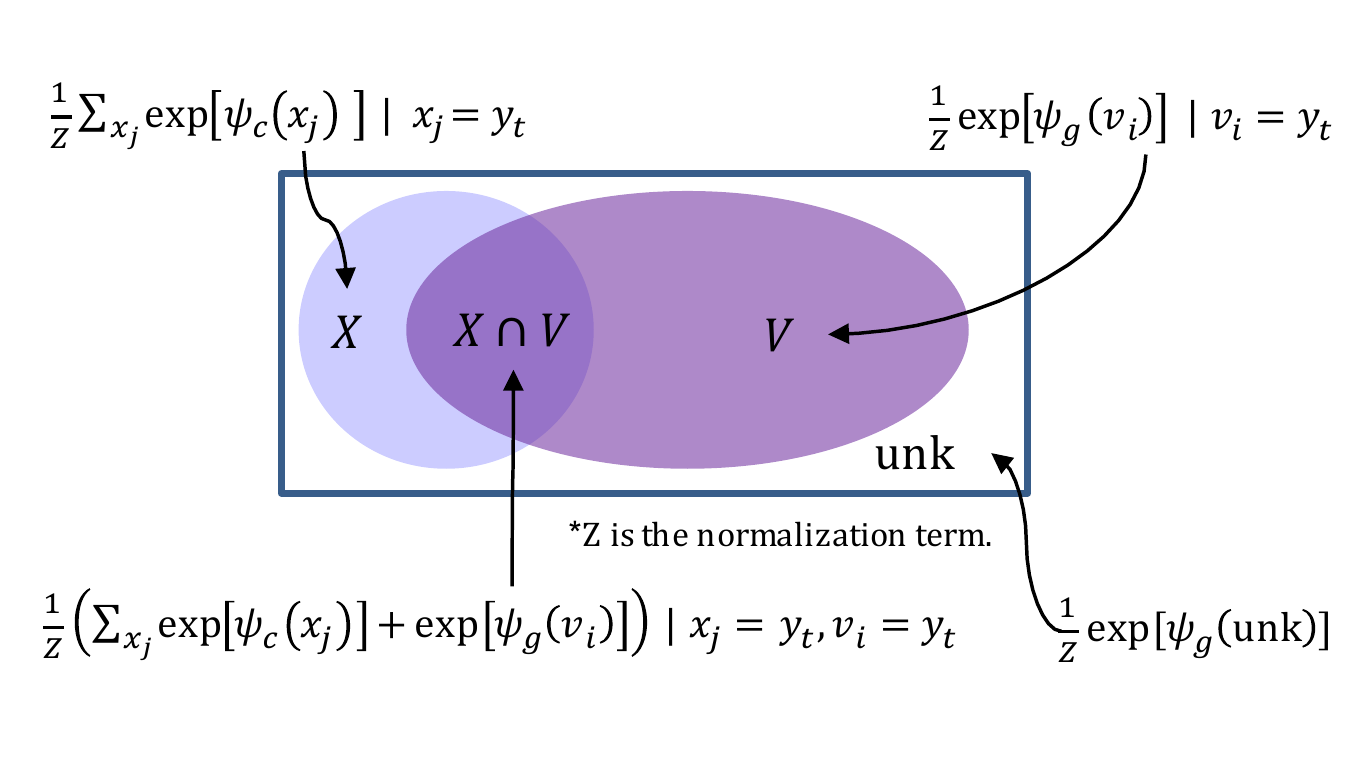} 
          	\vspace{-15pt}
          	\caption{\label{oov} The illustration of  the decoding probability $p(y_t|\cdot)$
          	 %which is seen 
          	 as a 4-class classifier. } 
   \vspace{-5pt}
  \end{figure}

%{\color{green} \textsc{CopyNet} predicts a word with a probabilistic model that mixes the generate-mode and the copy-mode, as illustrated in Figure~\ref{model}(b).}
Given the decoder RNN state $\mathbf{s}_t$ at time $t$ together with $\M$,      
%the encoded source $\h(\X) = \{\mathbf{h}_1, ..., \mathbf{h}_{T_S}\}$, 
the probability of generating any target word $y_t$,  is given by the ``mixture" of probabilities as follows
\begin{multline}
\hspace{-2pt}	p(y_t|\sss_t, y_{t-1}, \c_t, \M) = p(y_t, \textsf{\small g} | \sss_t, y_{t-1}, \c_t, \M)\\ + p(y_t, \textsf{\small c}|\sss_t, y_{t-1}, \c_t,\M) \label{eq:mix}
\end{multline}
where \textsf{\small g} stands for the generate-mode, and \textsf{\small c} the copy mode. The probability of the two modes 
are given respectively by
\begin{eqnarray} %\small 
\label{eq:pg}
p(y_t, \textsf{\small g}|\cdot) 
\hspace{-3pt } 
=&
\hspace{-11pt} 
 \left \{\begin{matrix}
\dfrac{1}{Z}e^{\psi_g(y_t)},                       \;\;\; \qquad& \;\;\; y_t \in \calV \\
0,                                                                          \;\;\; \qquad&\;\;\; y_t \in \calX \cap \bar{V}\\
\dfrac{1}{Z}e^{\psi_g(\textsc{unk})} \;\;\;\;     \;\;\; \qquad& \;\;\; y_t \not \in \calV \cup \calX     \\
\end{matrix}\right. \\
\label{eq:pc}
	p(y_t, \textsf{\small c}|\cdot)
	\hspace{-3pt } 
	=& 
	\hspace{-12pt} 
	\left \{\begin{matrix}
\dfrac{1}{Z}\sum_{j:x_j=y_t} e^{\psi_c(x_j)},  \hspace{-17pt} &y_t \in \calX \\
0 &\text{otherwise} 
\end{matrix}\right. 
\end{eqnarray}
\vspace{-5pt}
%\begin{equation} %\small 
%\label{eq:pg}
%p(y_t, \textsf{\small g}|\cdot) = \left \{\begin{matrix}
%e^{\psi_g(y_t)}/Z, ~ &y_t \in \calV \\
%0,~ & y_t \in \calX \cap \bar{V}\\
%e^{\psi_g(\textsc{unk})}/Z ~ & y_t \not \in \calV \cup \calX     \\
%\end{matrix}\right.
%\end{equation}
%\begin{equation}
%\label{eq:pc}
%	p(y_t, \textsf{\small c}|\cdot)\hspace{-3pt} = \hspace{-3pt} \left \{\begin{matrix}
%\sum_{j:x_j=y_t} e^{\psi_c(x_j)}/Z,  \hspace{-7pt} &y_t \in \calX \\
%0 &\text{otherwise}  \\
%\end{matrix}\right.
%\end{equation}
\\
where $\psi_g(\cdot)$ and $\psi_c(\cdot)$ are score functions for generate-mode and copy-mode, respectively, and $Z$ is the normalization term shared by the two modes, $Z = \sum_{v\in \cal V \cup \{\textsc{unk}\}}e^{\psi_g(v)} + \sum_{x\in X}e^{\psi_c(x)}.$
% [check the equation]. 
Due to the shared normalization term, the two modes are basically competing through a softmax function (see Figure~\ref{model} for an illustration with example), rendering Eq.(\ref{eq:mix}) different from the canonical definition of the mixture model~\cite{GaussianMixture}. This is also pictorially illustrated in Figure \ref{oov}. 
 The score of each mode is calculated:\vspace{-5pt}
\paragraph{Generate-Mode:}~The same scoring function as in the generic RNN encoder-decoder~\cite{bahdanau2014neural} is used, i.e.
\begin{equation}\label{eq:gen}
	\psi_g(y_t=v_i) = \vv_i^\top \W_o \sss_t, \quad v_i \in \calV \cup \textsc{unk}
\end{equation}
%\begin{equation}\label{eq:gen}
%	\psi_g(y_t) =\left \{\begin{matrix}v_i^TW_os_t, \quad &y_t = v_i \in V
%	\\ v_{\text{unk}}^TW_os_t, \quad &y_t \notin V
%	\end{matrix}\right.
%\end{equation}
where $\W_o \in \mathbb{R}^{(N+1) \times d_s}$ and $ \vv_i$ is the one-hot indicator vector for $v_i$. %We use the one-hot vectors in Eq.~\ref{eq:gen} for $\calV$ and $\textsc{unk}$. As you can see, symbols not in the vocabulary will all be regarded the same and assigned a fixed score.

\paragraph{Copy-Mode:}~The score for ``copying" the word $x_j$ is calculated as \vspace{-7pt}
\begin{equation}\label{eq:cp}
	\psi_c(y_t=x_j) =	\sigma\left(\mathbf{h}_j^\top \mathbf{W}_c\right)\mathbf{s}_t, \quad  x_j \in \calX \vspace{-7pt}
\end{equation}
%\begin{equation}\label{eq:cp}
%	\psi_c(y_t) =\left \{\begin{matrix}
%	 \sigma\left(h_j^TW_c\right)s_t, \quad &y_t = x_j \in X\\ 
%	-\infty, \quad &y_t \notin X
%	\end{matrix}\right.
%\end{equation}
where $\mathbf{W}_c\in \mathbb{R}^{d_h \times d_s}$, and $\sigma$ is a non-linear activation function, considering that the non-linear transformation in Eq.(~\ref{eq:cp}) can help project $s_t$ and $h_j$ in the same semantic space. Empirically,  we also found that using the $\tanh$ non-linearity worked better than linear transformation, and we used that for the following experiments.
When calculating the copy-mode score, we use the hidden states $\{\mathbf{h}_1, ..., \mathbf{h}_{T_S}\}$ to ``represent" each of the word in the source sequence $\{x_1, ..., x_{T_S}\}$ since the bi-directional RNN encodes not only the content, 
but also the location information into the hidden states in $\M$. The location informaton is important for copying (see Section \ref{sec:reading} for related discussion). 
Note that we sum the probabilities of all $x_j$ equal to $y_t$ in Eq.~(\ref{eq:pc}) considering that there may be multiple source symbols for decoding $y_t$. 
%\vspace{-5pt}
% where $y \in X \cup V \cup \{v_{\text{unk}}\}$ regarding all the symbols outside $X \cup V$ as the same symbol ``unk".
Naturally we let $p(y_t, \textsf{\small c}|\cdot)=0$ if $y_t$ does not appear in the source sequence, and set $p(y_t, \textsf{\small g}|\cdot)=0$ when $y_t$ only appears in the source. %Such changes make the probabilities sum up to 1.% and shrinks the range of the OOV words.

 \subsection{State Update}
\label{sec:stateupdate}
\textsc{CopyNet} updates each decoding state $\sss_{t}$ with the previous state $\sss_{t-1}$, the previous symbol $y_{t-1}$ and the context vector $\c_{t}$  following Eq.~(\ref{eq:dec}) for the generic attention-based Seq2Seq model. However, there is some minor changes in the $y_{t-1}\hspace{-3pt}\longrightarrow \hspace{-3pt}\sss_{t}$ path for the copying mechanism. More specifically, $y_{t-1}$ will be represented as $[\e({y_{t-1})} ; \zeta(y_{t-1})]^\top$, where $\e({y_{t-1}})$ is the word embedding associated with $y_{t-1}$, while $\zeta(y_{t-1})$ is the weighted sum of hidden states in $\M$ corresponding to $y_t$ \vspace{-6pt}
\begin{equation}
%\small
\label{eq:loc}
\begin{split}
%	&\zeta(y_{t-1}) = \left \{ \begin{matrix}
%		\sum\nolimits_{\tau=1}^{T_S}\rho_{t\tau} \h_{\tau} \quad & y_{t-1} \in \calX\\
%		\textbf{0} \quad & \text{otherwise} 
%	\end{matrix}\right.
	&\zeta(y_{t-1}) = \sum\nolimits_{\tau=1}^{T_S}\rho_{t\tau} \h_{\tau}\\
	&\rho_{t\tau} = \left \{ \begin{matrix} 
	     \dfrac{1}{K} p(x_\tau, \textsf{c} | \sss_{t-1}, \M),
	    %\dfrac{p(x_\tau, c | \sss_{t-1}, \h)}{\sum\nolimits_{\tau':x_{\tau'}=y_{t-1}} p(x_\tau', c | \sss_{t-1}, \h)}
	     \quad & x_{\tau} = y_{t-1}\\
		0 \quad & \text{otherwise} 
	\end{matrix}\right.
%	&\rho_{t\tau} = \text{Nor}\left[ p(x_\tau, c | \cdot) \odot \mathds{1}(y_{t-1} - x_{\tau})\right]\\
%	&\rho_{t\tau} = \hat{\rho}_{t\tau}/{\sum\nolimits_{\tau'}\hat{\rho}_{t\tau'}};  \hat{\rho}_{t\tau} = P(x_\tau) \cdot (y_t == x_{\tau})
\end{split} \vspace{-7pt}
\end{equation}
%\begin{equation}
%\label{eq:loc}
%\begin{split}
%	&\zeta(y_{t-1}) = \left \{ \begin{matrix}
%		\sum_{\tau=1}^{T_S}\rho_{t\tau} \h_{\tau} \quad & y_t \in X\\
%		\textbf{0} \quad & \text{otherwise} 
%	\end{matrix}\right.\\
%	&\rho_{t\tau} = \text{Nor}\left[ p(x_\tau, c | \cdot) \odot \mathds{1}(y_{t-1} - x_{\tau})\right]\\
%%	&\rho_{t\tau} = \hat{\rho}_{t\tau}/{\sum\nolimits_{\tau'}\hat{\rho}_{t\tau'}};  \hat{\rho}_{t\tau} = P(x_\tau) \cdot (y_t == x_{\tau})
%\end{split}
%\end{equation}
where $K$ is the normalization term which equals $ \sum_{{\tau'}:x_{\tau'}=y_{t-1}} p(x_{\tau'}, c | \sss_{t-1}, \M)$, considering there may exist multiple positions with $y_{t-1}$ in the source sequence. In practice, $\rho_{t\tau}$ is often concentrated on one location among multiple appearances, indicating the prediction is closely bounded to the location of words.

In a sense $\zeta(y_{t-1})$ performs a type of read to $\M$ similar to the attentive read (resulting $\c_t$) with however higher precision. In the remainder of this paper, $\zeta(y_{t-1})$ will be referred to as \emph{selective read}. $\zeta(y_{t-1})$ is specifically designed for the copy mode: with its pinpointing precision to the corresponding  $y_{t-1}$, it naturally bears the location of $y_{t-1}$ in the source sequence encoded in the hidden state. As will be discussed more in Section~\ref{sec:reading}, this particular design potentially helps copy-mode in covering a consecutive sub-sequence of words. If $y_{t-1}$ is not in the source,  we let $\zeta(y_{t-1})=\textbf{0}$.

%{\color{blue} In this paper $\zeta(y_{t-1})$ is named \emph{selective read}, since it can be viewed as another kind of read from the $\M$ when it fetches the hidden state for decoder. Different from attentive read, selective read at time $t$ is determined jointly by $y_{t}$ and $\sss_t$, which in practice is usually concentrated into one particular hidden state even if the word $y_{t}$ has multiple appearances in the source.
%}
 %Concatenating with the content-based input, we 
 
%{\color{blue} 
%The location-based input can be explained intuitively that in many tasks, we perform ``copy and paste" focusing more on the locations of the segments rather than the detailed content behind the them. Each segment typically consists of successive symbols, and it will be helpful to input the previous location to move on the target position. 
%}

\subsection{Hybrid Addressing of $\M$} 
\label{sec:reading}
We hypothesize that \textsc{CopyNet} uses a hybrid strategy for fetching the content in $\M$, which combines both content-based and location-based addressing. Both addressing strategies are coordinated by the decoder RNN in managing the attentive read and selective read, as well as determining when to enter/quit the copy-mode. 

Both the semantics of a word and its location in $X$ will be encoded into the hidden states in $\M$ by a properly trained encoder RNN. Judging from our experiments, the attentive read of \textsc{CopyNet} is driven more by the semantics and language model, therefore capable of traveling more freely on $\M$, even across a long distance. On the other hand, once \textsc{CopyNet} enters the copy-mode, the selective read of $\M$ is often guided by the location information. As the result, the selective read often takes rigid move and tends to cover consecutive words, including \textsc{unk}s.  Unlike the explicit design for hybrid addressing in Neural Turing Machine~\cite{graves2014neural,kurach2015neural}, \textsc{CopyNet} is more subtle: it provides the architecture that can facilitate some particular location-based addressing and lets the model figure out the details from the training data for specific tasks. \\\vspace{-7pt} \\
\textbf{Location-based Addressing:} 
With the location information in $\{\h_i\}$, the information flow  \vspace{-5pt} %(need to define earlier) 
\[
%\cdots \rightarrow
\zeta(y_{t-1}) \xrightarrow{\text{update}}  \mathbf{s}_{t} \xrightarrow{\text{predict}} y_{t} \xrightarrow{\text{sel. read}}  \zeta(y_{t}) \vspace{-5pt} 
%\rightarrow \cdots 
\]
provides a simple way of ``moving one step to the right" on $X$. More specifically, assuming the selective read $\zeta(y_{t-1})$ concentrates on the $\ell^{th}$ word in $X$, the state-update operation $\zeta(y_{t-1})\hspace{-4pt} \xrightarrow{\text{update}} \hspace{-4pt} \mathbf{s}_t $ acts as ``\texttt{\small location} $\leftarrow$ \texttt{\small location+1}", making $\sss_t$ favor the $(\ell \hspace{-3pt}+\hspace{-3pt}1)^{th}$ word in $X$ in the prediction $\mathbf{s}_t \xrightarrow{\text{predict}} y_{t}$ in copy-mode. This again leads to the selective read  $\hat{h}_{t} \hspace{-3pt}\xrightarrow{\text{sel. read}} \hspace{-3pt} \zeta(y_{t})$ for the state update of the next round.
 \\
 \vspace{-7pt}\\
\textbf{Handling Out-of-Vocabulary Words}~
Although it is hard to verify the exact addressing strategy as above directly, there is strong evidence from our empirical study. Most saliently, a properly trained \textsc{CopyNet} can copy a fairly long segment full of OOV words, despite the lack of semantic information in its $\M$ representation. This provides a natural way to extend the effective vocabulary to include all the words in the source. Although this change is small, it seems quite significant empirically in alleviating the OOV problem. Indeed, for many NLP applications (e.g., text summarization or spoken dialogue system),  much of the OOV words on the target side, for example the proper nouns, are essentially the replicates of those on the source side.

\vspace{-5pt}

\section{Learning} 	\vspace{-5pt}
Although the copying mechanism uses the ``hard" operation to copy from the source and choose to paste them or generate symbols from the vocabulary, \textsc{CopyNet} is fully differentiable and can be optimized in an end-to-end fashion using back-propagation. Given the batches of the source and target sequence $\{X\}_N$ and $\{Y\}_N$, the objectives are to minimize the negative log-likelihood:
	\vspace{-9pt}\begin{equation}
      \hspace{-0.5pt} \mathcal{L} = -\frac{1}{N}\sum_{k=1}^N\sum_{t=1}^{T}\log \left[ p(y^{(k)}_t|y^{(k)}_{<t}, X^{(k)})\right], 	\vspace{-9pt}
%     \mathcal{L} = -\frac{1}{N}\sum_{k=1}^N\sum_{t=1}^{T}\log \left[ P(y^{(k)}_t, g|\cdot) + P(y^{(k)}_t, c|\cdot)\right]
\end{equation}
where we use superscripts to index the instances. Since the probabilistic model for observing any target word is a mixture of generate-mode and copy-mode, there is no need for any additional labels for modes. The network can learn to coordinate the two modes from data. More specifically, if one particular word $y_{t}^{(k)}$ can be found in the source sequence, the copy-mode will contribute to the mixture model, and the gradient will more or less encourage the copy-mode; otherwise, the copy-mode is discouraged due to the competition from the shared normalization term $Z$. In practice, in most cases one mode dominates.
%Although the copying mechanism uses the ``hard" operation to copy from the source and choose to paste them or generate symbols from the vocabulary, \textsc{CopyNet} is fully differentiable and can be optimized in an end-to-end fashion using back-propagation. Given the batches of the source and target sequence $\{X\}_N$ and $\{Y\}_N$,  our objective is to minimize the negative log-likelihood:
%\begin{equation}
%	\mathcal{L} = -\frac{1}{N}\sum_{k=1}^N\sum_{t=1}^{T}\log \left[ p(y^{(k)}_t|y^{(k)}_{<t}, X^{(k)})\right]
%%	\mathcal{L} = -\frac{1}{N}\sum_{k=1}^N\sum_{t=1}^{T}\log \left[ P(y^{(k)}_t, g|\cdot) + P(y^{(k)}_t, c|\cdot)\right]
%\end{equation}
%where we directly optimize \textsc{CopyNet} over the mixture distribution for at least two reasons:
%\begin{itemize}
%	\vspace{-8pt}
%	\item Learning two modes simultaneously avoids making hard decisions on switching modes.
%	\vspace{-18pt}
%	\item No additional labels for mode switching is  required. The network can learn to switch and make full use of the two modes automatically.
%\end{itemize}
% Each decoded symbol is seen as a mixture of the generate mode and copy mode, although in most cases one mode will naturally dominate the other.
%\\ \vspace{-7pt} \\
%\textbf{How CopyNet learns to ``copy" correctly ?}\\
%Brief analysis about that.  "Content Triggers ?"
%In practise, to encourage using the copy mode and avoid decoding the OOVs, 

\section{Experiments}
We report our empirical study of \textsc{CopyNet} on the following three tasks with different characteristics
\begin{enumerate}
	\vspace{-7pt}
	\item A synthetic dataset on with simple patterns;
	\vspace{-7pt}
	\item A real-world task on text summarization;
	\vspace{-7pt}
	\item A dataset for simple single-turn dialogues.
\end{enumerate}  

\subsection{Synthetic Dataset}
\label{section: synthetic}
\textbf{Dataset:}~  
We first randomly generate transformation rules with 5$\sim$20 symbols and variables $\mathbf{x}$ \& $\mathbf{y}$, e.g.  \vspace{-7pt} 
\[
\texttt{a b } \mathbf{x} \texttt{ c d }\mathbf{y} \texttt{ e f} \longrightarrow \texttt{ g h }\mathbf{x}\texttt{ m},  
\vspace{-5pt}
\]
with \{\texttt{a\;b\;c\;d\;e\;f\;g\;h\;m}\} being regular symbols from a vocabulary of size 1,000. As shown in the table below, each rule can further produce a number of instances by replacing the variables with randomly generated subsequences (1$\sim$15 symbols) from the same vocabulary. We create five types of rules, including ``$\mathbf{x}\rightarrow$ $\emptyset$".
The task is to learn to do the Seq2Seq transformation from the training instances. 
This dataset is designed to study the behavior of \textsc{CopyNet} on handling simple and rigid patterns. Since the strings to repeat are random, they can also be viewed as some extreme cases of rote memorization.   
%{\color{red}This dataset is designed to demonstrate the clear mechanism of \textsc{CopyNet}. }
%, instances generated are designed to evaluate the copying mechanism in each situation.
% We create the synthetic dataset named ``Simple Rules for \textsc{Seq2seq} Learning (SRSS)" for this analysis.  
%The dataset contains different symbols with a vocabulary of 1,000 words (with indices from $0$ to $999$) and two variables \textbf{X} \& \textbf{Y}.  
%%%%%%%%%%%%%%%%%%%%%%%%%%%%%%%%%%%%%%%%%%%%%%%%
 \begin{table}[!ht] % "[h!]" location specifier just for this example
\small
\centering
\begin{tabular}{l|l}
\toprule
 Rule-type& {Examples (e.g. $\mathbf{x}$ = $\texttt{i h k}$,~~ $\mathbf{y}$ = $\texttt{j c}$)} \\
\midrule
$\mathbf{x}$ $\rightarrow \emptyset$ 
& $\texttt{a b c d }\mathbf{x} \texttt{ e f} \rightarrow \texttt{c d g}$\\
\cmidrule{1-2}
$\mathbf{x}$ $\rightarrow$ $\mathbf{x}$ 
& $\texttt{a b c d }\mathbf{x} \texttt{ e f} \rightarrow \texttt{c d } \mathbf{x} \texttt{ g} $\\
$\mathbf{x}$ $\rightarrow$ $\mathbf{x\,x}$ 
& $\texttt{a b c d }\mathbf{x} \texttt{ e f} \rightarrow \mathbf{x}  \texttt{ d } \mathbf{x} \texttt{ g} $\\
$\mathbf{x\,y}$ $\rightarrow$ $\mathbf{x}$ 
& $\texttt{a b } \mathbf{y} \texttt{ d }\mathbf{x} \texttt{ e f} \rightarrow \mathbf{x}  \texttt{ d i g} $\\
$\mathbf{x\,y}$ $\rightarrow$ $\mathbf{x\,y}$ 
& $\texttt{a b } \mathbf{y} \texttt{ d }\mathbf{x} \texttt{ e f} \rightarrow \mathbf{x}  \texttt{ d } \mathbf{y} \texttt{ g} $\\
\bottomrule
\end{tabular} \vspace{-5pt}
\end{table} \vspace{-2pt}\\ \vspace{-15pt}\\%%%%%%%%%%%%%%%%%%%%%%%%%%%%%%%%%%%%%%%%%%%%%%%%
\textbf{Experimental Setting:}~We select 200 artificial rules from the dataset, and for each rule 200 instances are generated, which will be split into training (50\%) and testing (50\%).  We compare the  accuracy of \textsc{CopyNet} and the RNN Encoder-Decoder with (i.e. RNNsearch) or without attention (denoted as Enc-Dec).
For a fair comparison, we use bi-directional GRU for encoder and another GRU for decoder  for all Seq2Seq models, with hidden layer size = 300 and word embedding dimension = 150. We use bin size = 10 in beam search for testing. The prediction is considered correct only when the generated sequence is exactly the same as the given one. 
%It is clear that the system must make full use of the limited instances to summarize patterns and then output the results correctly.
%\\ \vspace{-7pt} \\

%We compare the accuracy of \textsc{CopyNet} and the baseline RNNSearch. It is clear from Table~\ref{table-acc} that \textsc{CopyNet} significantly outperforms RNNsearch on all rule-types except ``$\mathbf{x} \rightarrow \emptyset$", indicating that \textsc{CopyNet} can effectively learn the patterns from instances and accurately repeat rather long subsequence of symbols at the proper places. {\color{blue} This is often hard to an canonical encoder-decoder framework due to the difficulty of representing a long sequence with very high fidelity. This difficulty can be alleviated with the attention mechanism,

It is clear from Table~\ref{table-acc} that \textsc{CopyNet} significantly outperforms the other two on all rule-types except ``$\mathbf{x} \rightarrow \emptyset$", indicating that \textsc{CopyNet} can effectively learn the patterns with variables and accurately replicate rather long subsequence of symbols at the proper places.This is hard to Enc-Dec due to the difficulty of representing a long sequence with very high fidelity. This difficulty can be alleviated with the attention mechanism. However attention alone seems inadequate for handling the case where strict replication is needed. 
%{\color{red}but ...  A closer look at \textsc{CopyNet} trained on this data set reveals that ... }
%%%%%%%%%%%%%%%%%%%%%%%%%%%%%%%%%%%%%%%%%%%%%%%%
 \begin{table}[tb] % "[h!]" location specifier just for this example
\small
\centering
\begin{tabular}{lccccc}
\toprule
Rule-type& $\mathbf{x}$ &  $\mathbf{x}$ &   $\mathbf{x}$ &   $\mathbf{xy}$&    $\mathbf{xy}$ \\% & Avg. \\
&  $\rightarrow \emptyset$ &  $\rightarrow\mathbf{x}$ &  $\rightarrow\mathbf{xx}$ &  $\rightarrow\mathbf{x}$&   $\rightarrow\mathbf{xy}$  \\
\midrule
%RNNEncDec
%& 110  &&&&& \\
%\cmidrule{1-2}
Enc-Dec
& \textbf{100}  & 3.3 & 1.5 & 2.9 & 0.0\\
RNNSearch 
& 99.0  & 69.4 & 22.3 & 40.7 & 2.6\\
\midrule
\textsc{CopyNet}
& 97.3  & \textbf{93.7} & \textbf{98.3} & \textbf{68.2} & \textbf{77.5}\\
\bottomrule
\end{tabular} \vspace{-5pt}
\caption{\label{table-acc}  The test accuracy (\%) on synthetic data.}
%The decoding accuracy in percentage on the testing set with different rule-types.} % Three kinds of instances are evaluated for our model: from top to bottom are (a) repeated  (b) ordered, and (c) random, respectively.}
\vspace{-16pt}
\end{table} %\vspace{-15pt}
%%%%%%%%%%%%%%%%%%%%%%%%%%%%%%%%%%%%%%%%%%%%%%%%
%\textbf{How \textsc{CopyNet} learns to copy?}\\
%One advantage of a synthetic dataset is that it helps to explain the mechanism when copying.  Seeing Figure~\ref{syn}, we visualize the 
%%values of $P(y_t, c|s_t, \bm{h})$ 
%decoding process of one transformation for a better understanding to what happens when the copying mechanism works.

\begin{figure}[h!]
   	\centering
          	\vspace{-5pt}
          	\includegraphics[width=.98\linewidth]{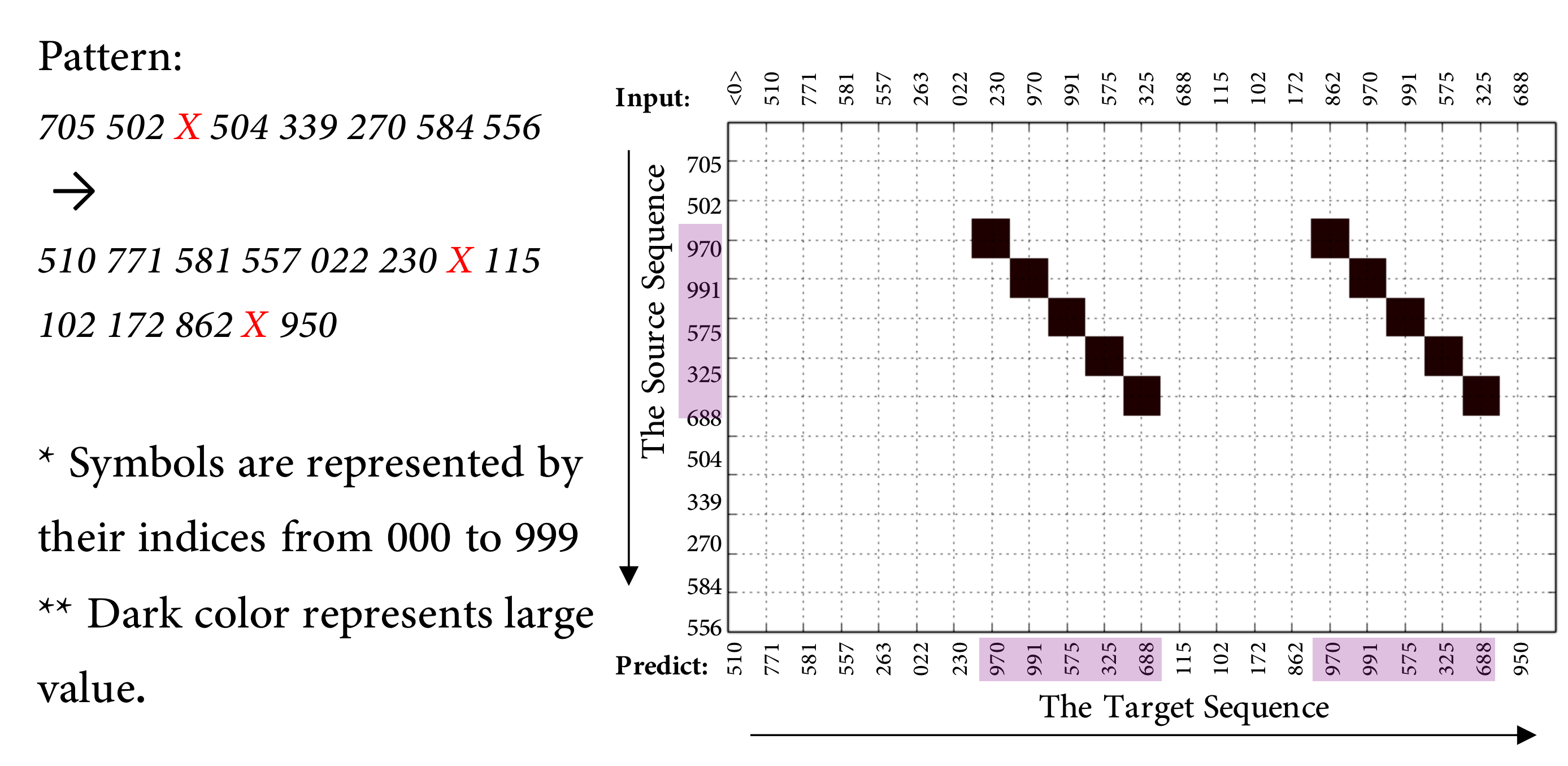} 
          	\vspace{-10pt}
          	\caption{\label{syn} Example output of \textsc{CopyNet} on the synthetic dataset. The heatmap represents the activations of the copy-mode over the input sequence (left) during the decoding process (bottom).} 
   \vspace{-17pt}
  \end{figure}

% It is clear that the copy mode is activated only when copying substrings, and stays zero while generating symbols. Seeing the illustrated gate activations, firstly the decoder of \textsc{CopyNet} learns to copy the first symbol of substring 
% %based on the content information from the encoder. It 
% and turns off the update gates to avoid copied symbols read into the hidden states to keep the copying moving on smoothly until the last symbol. Such mechanism partially explains why \textsc{CopyNet} can efficiently copy arbitrarily strings without affecting the performance.  
%
A closer look (see Figure~\ref{syn} for example) reveals that the decoder is dominated by copy-mode when moving into the subsequence to replicate, and switch to generate-mode after leaving this area, showing \textsc{CopyNet} can achieve a rather precise coordination of the two modes. \vspace{-5pt}    

\subsection{Text Summarization}
Automatic text summarization aims to find a condensed representation which can capture the core meaning of the original document. It has been recently formulated as a Seq2Seq learning problem in~\cite{rush2015neural,hu2015lcsts}, which essentially gives \emph{abstractive} summarization since the summary is generated based on a representation of the document. In contrast, \textit{extractive} summarization extracts sentences or phrases from the original text to fuse them into the summaries, therefore making better use of the overall structure of the original document. In a sense, \textsc{CopyNet} for summarization lies somewhere between two categories, since part of output summary is actually extracted from the document (via the copying mechanism), which are fused together possibly with the words from the generate-mode.\vspace{-5pt}

%%As mentioned above,  encoder-decoder models easily come across the problems of OOVs when doing summarization. 
%However,
%%have to encode the whole input text for ``understanding" their meanings implicitly to generate a summary, which sometimes is not essential and comes across the problem of OOVs. 
%on the perspectives of  \textsc{CopyNet},  its generate and copy modes just correspond to the \textit{abstractive} and \textit{extractive} behaviours in summarization, respectively. \\ \vspace{-7pt} \\
\paragraph{Dataset:}~We evaluate our model on the recently published LCSTS dataset~\cite{hu2015lcsts},  a large scale dataset for short text summarization. The dataset is collected from the news medias on Sina Weibo\footnote{www.sina.com} including pairs of (short news, summary) in Chinese. Shown in Table~\ref{table-lcsts},  PART \uppercase\expandafter{\romannumeral2}  and  \uppercase\expandafter{\romannumeral3} are manually rated for their quality from 1 to 5. Following the setting of~\cite{hu2015lcsts} we use Part \uppercase\expandafter{\romannumeral1}  as the training set and and the subset of Part  \uppercase\expandafter{\romannumeral3} scored from 3 to 5 as the testing set. \vspace{-5pt}
  %%%%%%%%%%%%%%%%%%%%%%%%%%%%%%%%%%%%%%%%%%%%%%%%
 \begin{table}[!ht] % "[h!]" location specifier just for this example
\small
\centering
\begin{tabular}{l|ccc}
\toprule
Dataset & PART \uppercase\expandafter{\romannumeral1}  & PART \uppercase\expandafter{\romannumeral2}  & PART \uppercase\expandafter{\romannumeral3} \\
\midrule
no. of pairs & 2,400,591  &  10,666 & 1106\\
no. of $\text{score} \geq 3$ & - & 8685 & 725 \\
\bottomrule
\end{tabular} \vspace{-5pt}
\caption{\label{table-lcsts} Some statistics of the LCSTS dataset.} % Three kinds of instances are evaluated for our model: from top to bottom are (a) repeated  (b) ordered, and (c) random, respectively.}
 \vspace{-8pt}
\end{table}%\\
%%%%%%%%%%%%%%%%%%%%%%%%%%%%%%%%%%%%%%%%%%%%%%%%
\\\textbf{Experimental Setting:}~We try \textsc{CopyNet} that is based on character (+C) and word (+W). For the word-based variant the word-segmentation is obtained with jieba\footnote{https://pypi.python.org/pypi/jieba}. We set the vocabulary size to 3,000 (+C) and 10,000 (+W) respectively, which are much smaller than those for models in~\cite{hu2015lcsts}. For both variants we set the embedding dimension to 350 and the size of hidden layers to 500. %\\ \vspace{-7pt} \\
%\textsc{CopyNet} is optimized on the training set through Adam~\cite{kingma2014adam} using a batch size of 20. 
Following~\cite{hu2015lcsts}, we evaluate the test performance with the commonly used ROUGE-1, ROUGE-2 and ROUGE-L~\cite{lin:2004:ACLsummarization}, and compare it against the two models in \cite{hu2015lcsts}, which are essentially canonical Encoder-Decoder and its variant with attention.  \vspace{-5pt}
 %%%%%%%%%%%%%%%%%%%%%%%%%%%%%%%%%%%%%%%%%%%%%%%%
%We need to generate synthetic sequences with suitable patterns for \textsc{Seq2seq} learning.
 \begin{table}[htb] % "[h!]" location specifier just for this example
\small
%\vspace{-20pt}
\centering
\begin{tabular}{llccc}
\toprule
 Models&& \multicolumn{3}{c}{ROUGE scores on LCSTS (\%)} \\
%\cmidrule{2-3}
 && R-1 &\hspace{14pt} R-2 & R-L	\\
\midrule
RNN &  +C  & 21.5 & \hspace{14pt}8.9 & 18.6 \\
\cite{hu2015lcsts} & +W & 17.7 & \hspace{14pt}8.5 & 15.8 \\
%\cmidrule{3-5}
RNN context   &  +C & 29.9 & \hspace{14pt}17.4 & 27.2 \\
\cite{hu2015lcsts} & +W & 26.8 & \hspace{14pt}16.1 & 24.1 \\
%\cmidrule{3-5}
\midrule
\multirow{2}{*}{\textsc{CopyNet}}& +C & \textbf{34.4} & \hspace{14pt}\textbf{21.6} & \textbf{31.3} \\
												    & +W & \textbf{35.0} & \hspace{14pt}\textbf{22.3} & \textbf{32.0} \\												  
\bottomrule
\end{tabular} \vspace{-5pt}
\caption{\label{table-summary} Testing performance of LCSTS, where ``RNN" is canonical Enc-Dec, and ``RNN context" its attentive variant.}
\vspace{-8pt}
\end{table} 
% \begin{table}[!ht] % "[h!]" location specifier just for this example
%\small
%\centering
%\begin{tabular}{lcccccc}
%\toprule
% Models& \multicolumn{6}{c}{LCSTS} \\
%%\cmidrule{2-3}
%& \multicolumn{2}{c}{ROUGE-1}  &\multicolumn{2}{c}{ROUGE-2} &\multicolumn{2}{c}{ROUGE-L}	\\
%\midrule
%EncDec & 10.11 & 10.11 & 10.11& 10.11 & 10.11 & 10.11 \\
%\bottomrule
%\end{tabular} \vspace{-5pt}
%\caption{\label{table-exp} Evaluation results.}
%\end{table} \vspace{-7pt}
%%%%%%%%%%%%%%%%%%%%%%%%%%%%%%%%%%%%%%%%%%%%%%%%%

\vspace{-5pt}  
It is clear from Table~\ref{table-summary} that \textsc{CopyNet} beats the competitor models with big margin. \newcite{hu2015lcsts} reports that the performance of a word-based model is inferior to a character-based one. One possible explanation is that a word-based model, even with a much larger vocabulary (50,000 words in \newcite{hu2015lcsts}), still has a large proportion of OOVs due to the large number of entity names in the summary data and the mistakes in word segmentation. \textsc{CopyNet}, with its ability to handle the OOV words with the copying mechanism, performs however slightly better with the word-based variant.
%, \textsc{CopyNet} combines the \textit{abstractive} and \textit{extractive} behaviours in one model, naturally solving the OOV problems by extracting them directly through the copy mode. The results show that the performance of the word-based CopyNet works slightly better than the character-based model, expanding the advantages of using words.
% \vspace{-10pt} 
 \begin{figure*}[tpb]
   	\centering
          	\vspace{-10pt}
          	\includegraphics[width=0.99\linewidth]{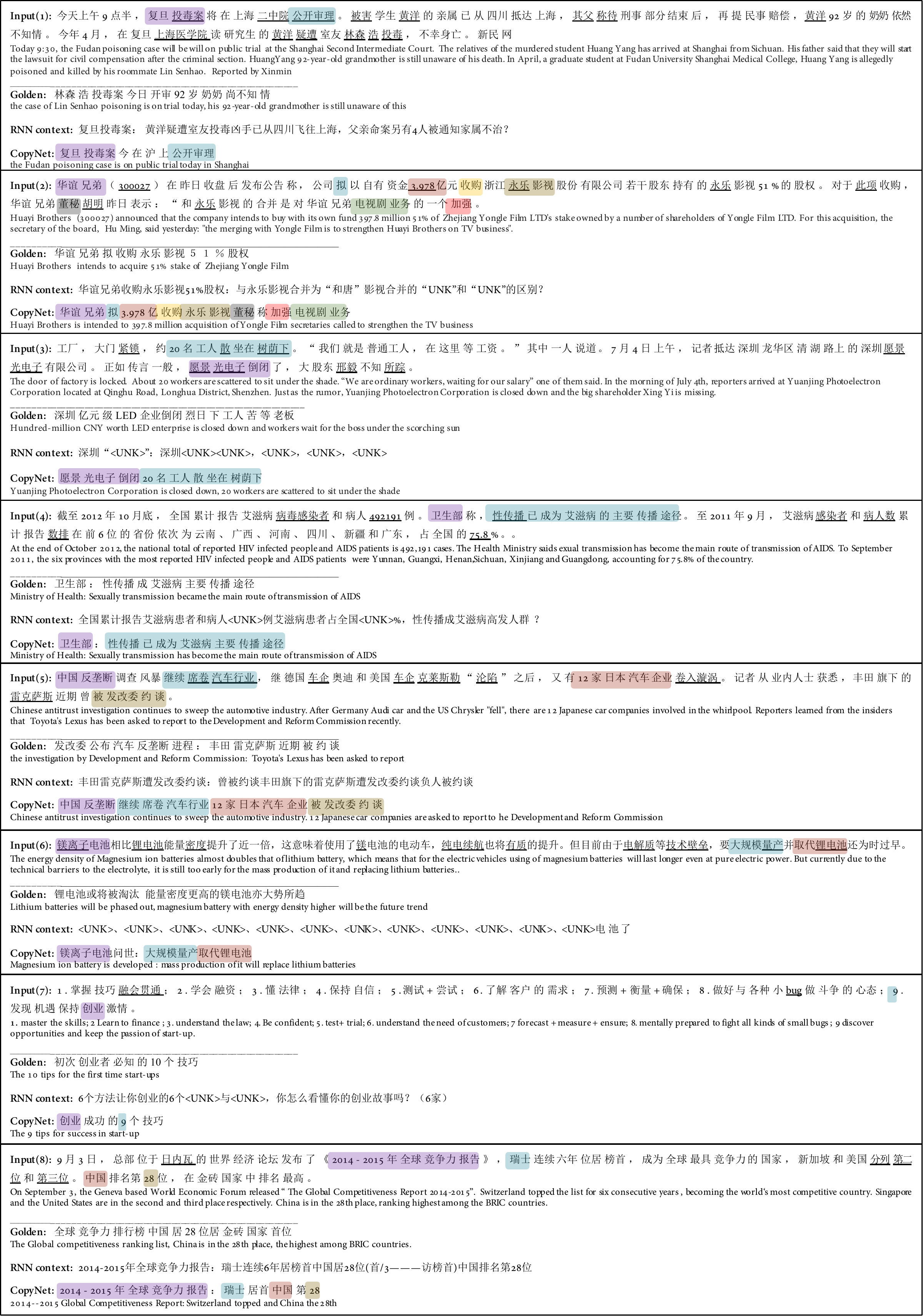}  
          	\vspace{-5pt}
          	\caption{\label{summary} Examples of \textsc{CopyNet} on LCSTS compared with RNN context. Word segmentation is applied on the input, where OOV words are underlined. The highlighted words (with different colors) are those words with copy-mode probability higher than the generate-mode.  We also provide literal English translation for the document, the golden, and \textsc{CopyNet}, while omitting that for RNN context since the language is broken.} 
   \vspace{-12pt}
  \end{figure*}   

 \subsubsection{Case Study}
As shown in Figure~\ref{summary},  we make the following interesting observations about the summary from \textsc{CopyNet}: 1) most words  are from copy-mode, but the summary is usually still fluent; 2) \textsc{CopyNet} tends to cover consecutive words in the original document, but it often puts together segments far away from each other, indicating a sophisticated coordination of content-based addressing and location-based addressing; 3) \textsc{CopyNet} handles OOV words really well: it can generate acceptable summary for document with many OOVs, and even the summary itself often contains many OOV words. In contrast, the canonical RNN-based approaches often fail in such cases.

It is quite intriguing that \textsc{CopyNet} can often find important parts of the document, a behavior with the characteristics of extractive summarization, while it often  generate words to ``connect" those words, showing its aspect of abstractive summarization. %This makes it an elegant and powerful hybrid. 

%Due to the nature of task, \textsc{CopyNet} often generate good summary but still vastly different the golden. Conversely, without a copying mechanism, RNNSearch works abnormally on complex input documents, especially when there exist many OOV words. Instead as shown in these examples, \textsc{CopyNet} directly handles the OOV problem by copying without any bad affects.  
%The copied segments are separately distributed in the source document, and \textsc{CopyNet} naturally learns to distinguish the important parts and extract them. We also find that  generate modes are sometimes activated to connect the extracted segments. Such behavior is quite similar with human when doing \textit{extractive} and \textit{abstractive} summarization together. 

\subsection{Single-turn Dialogue}
In this experiment we follow the work on neural dialogue model proposed in ~\cite{shang2015neural,vinyals2015neural,sordoni2015neural}, and test \textsc{CopyNet} on single-turn dialogue. Basically, the neural model learns to generate a response to user's input, from the given (input, response) pairs as training instances.
\\ \vspace{-7pt} \\
\textbf{Dataset:}~ 
%Building a suitable real-world dataset for dialogue system is difficult since it contains too many noisy patterns, spoken language and external knowledge, which influences the demonstration of our model.  
 %As shown in Figure~\ref{ds}, 
 We build a simple dialogue dataset based on the following three instructions: 
%  \begin{figure}[htbp]
%   	\centering
%          	\vspace{-3pt}
%          	\includegraphics[width=.98\linewidth]{ds.pdf} 
%          	\vspace{-5pt}
%          	\caption{\label{ds} Three steps to build the dialogue dataset} 
%   \vspace{-5pt}
%  \end{figure}   
  \begin{figure*}[htbp]
   	\centering
          	\vspace{-3pt}
          	\includegraphics[width=.99\linewidth]{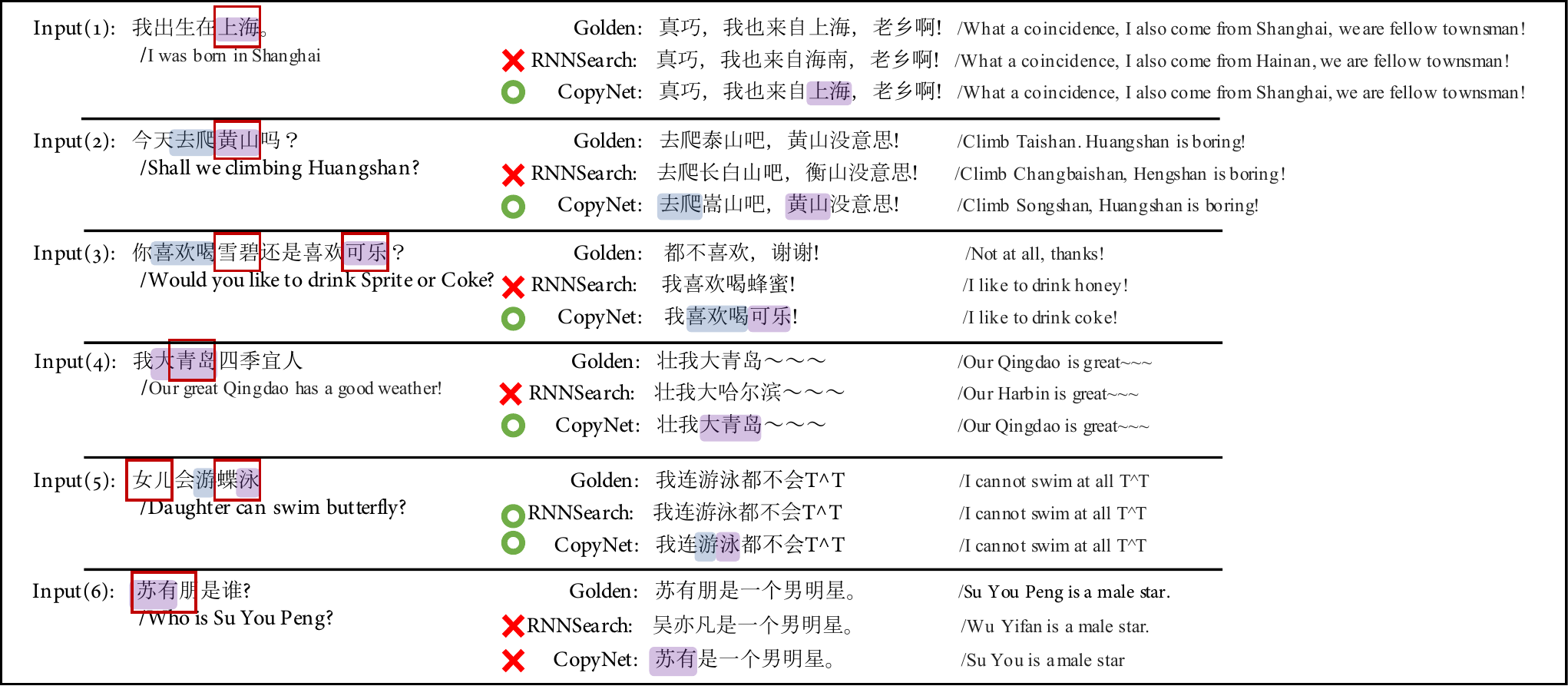} 
          	\vspace{-5pt}
          	\caption{\label{ds-exp} Examples from the testing set of DS-II shown as the input text and golden, with the outputs of RNNSearch and CopyNet. Words in red rectangles are unseen in the training set. The highlighted words (with different colors) are those words with copy-mode probability higher than the generate-mode. Green cirles (meaning correct) and red cross (meaning incorrect) are given based on human judgment on whether the response is appropriate. } 
   \vspace{-12pt}
  \end{figure*}   
\begin{enumerate}
	\vspace{-5pt}
	\item Dialogue instances are collected from Baidu Tieba\footnote{http://tieba.baidu.com} with some coverage of conversations of real life， e.g., greeting and sports, etc.\vspace{-7pt}
	\item Patterns with slots like 	\vspace{-10pt}
	\[\texttt{hi, my name is } \mathbf{x}\rightarrow \texttt{hi, } \mathbf{x} 	\vspace{-10pt} \]   
	are mined from the set, with possibly multiple responding patterns to one input.
	\vspace{-2pt}
	\item Similar with the synthetic dataset, we enlarge the dataset by filling the slots with suitable subsequence (e.g. name entities, dates, etc.)
	\vspace{-5pt}
\end{enumerate} 
%When filtering the retrieved answers, we focus on the answers that strongly related with the content of the posts, where a large proportion of answers contain the same substring with the input. Note that, 
To make the dataset close to the real conversations, we also maintain a certain proportion of instances with the response that 1) do not contain entities or 2) contain entities not in the input.  \\ \vspace{-7pt}\\
\textbf{Experimental Setting:} We create two datasets: DS-\uppercase\expandafter{\romannumeral1} and DS-\uppercase\expandafter{\romannumeral2} with slot filling on 173 collected patterns. The main difference between the two datasets is that the filled substrings for training and testing in DS-\uppercase\expandafter{\romannumeral2} have no overlaps, while in DS-\uppercase\expandafter{\romannumeral1} they are sampled from the same pool. For each dataset we use 6,500 instances for training and 1,500 for testing. We compare \textsc{CopyNet} with canonical RNNSearch, both character-based, with the same model configuration in Section~\ref{section: synthetic}. \vspace{-5pt} 

%%%%%%%%%%%%%%%%%%%%%%%%%%%%%%%%%%%%%%%%%%%%%%%%
\begin{table}[!ht] % "[h!]" location specifier just for this example
\small
\centering
\begin{tabular}{lccccc}
\toprule
 &\multicolumn{2}{c}{DS-\uppercase\expandafter{\romannumeral1}~(\%)}
 && \multicolumn{2}{c}{DS-\uppercase\expandafter{\romannumeral2}~(\%)}\\
%\cmidrule{2-3}
\cmidrule{2-3}\cmidrule{5-6}
Models &Top1&Top10 &&Top1 &Top10	\\
\midrule
RNNSearch &  44.1 & 57.7 && 13.5 & 15.9 \\
\textsc{CopyNet} & \textbf{61.2} & \textbf{71.0}  && \textbf{50.5} & \textbf{64.8} \\										  
\bottomrule
\end{tabular} \vspace{-5pt}
\caption{\label{table-ds} The decoding accuracy on the two testing sets. Decoding is admitted success only when the answer is found exactly in the Top-K outputs. }
\vspace{-8pt}
\end{table} 
 %%%%%%%%%%%%%%%%%%%%%%%%%%%%%%%%%%%%%%%%%%%%%%%%
We compare \textsc{CopyNet} and RNNSearch on DS-\uppercase\expandafter{\romannumeral1} and  DS-\uppercase\expandafter{\romannumeral2} in terms of top-1 and top-10 accuracy (shown in Table~\ref{table-ds}), estimating respectively the chance of the top-1 or one of top-10 (from beam search) matching the golden. Since there are often many good responses to an input, top-10 accuracy appears to be closer to the real world setting. 

As shown in Table~\ref{table-ds}, \textsc{CopyNet} significantly  outperforms RNNsearch, especially on DS-\uppercase\expandafter{\romannumeral2}. It suggests that introducing the copying mechanism helps the dialogue system master the patterns in dialogue and correctly identify the correct parts of input, often proper nouns, to replicate in the response. Since the filled substrings have no overlaps in DS-\uppercase\expandafter{\romannumeral2}, the performance of RNNSearch drops significantly as it cannot handle words unseen in training data. In contrast, the performance of \textsc{CopyNet} only drops slightly as it has learned to fill the slots with the copying mechanism and relies less on the representation of the words.

%\\\vspace{-7pt} \\
\subsubsection{Case Study} As indicated by the examples in  Figure~\ref{ds-exp}, \textsc{CopyNet} accurately replicates the critical segments from the input with the copy-mode, and generates the rest of the answers smoothly by the generate-mode. Note that in (2) and (3), the decoding sequence is not exactly the same with the standard one, yet still correct regarding to their meanings. In contrast, although RNNSearch usually generates answers in the right formats, it fails to catch the critical entities in all three cases because of the difficulty brought by the unseen words.

\vspace{-5pt}
\section{Related Work}
\vspace{-5pt}
Our work is partially inspired by the recent work of Pointer Networks~\cite{vinyals2015pointer}, in which a pointer mechanism (quite similar with the proposed copying mechanism) is used to predict the output sequence directly from the input. In addition to the difference with ours in application, \cite{vinyals2015pointer} cannot predict outside of the set of input sequence, while \textsc{CopyNet} can naturally combine generating and copying.  

\textsc{CopyNet} is also related to the effort to solve the OOV problem in neural machine translation. \newcite{luong-EtAl:2015:ACL-IJCNLP} introduced a heuristics to post-process the translated sentence using annotations on the source sentence. In contrast \textsc{CopyNet} addresses the OOV problem in a more systemic way with an end-to-end model. However, as \textsc{CopyNet} copies the exact source words as the output, it cannot be directly applied to machine translation. However, such copying mechanism can be naturally extended to any types of references except for the input sequence, which will help in applications with heterogeneous source and target sequences such as machine translation.

The copying mechanism can also be viewed as carrying information over to the next stage without any nonlinear transformation.  Similar ideas are proposed for training very deep neural networks in \cite{srivastava2015highway,he2015deep} for classification tasks, where shortcuts are built between layers for the direct carrying of information. % Copying from the source sequence can also be seen as a special effort to utilise "memory", as we discussed as a hybrid representation of short-term memory in Seq2Seq learning framework, sharing some similarity with memory-based approaches like \cite{weston2014memory,sukhbaatar2015end,graves2014neural}

Recently, we noticed some parallel efforts towards modeling mechanisms similar to or related to copying. \newcite{Cheng2016Neural} devised a neural summarization model with the ability to extract words/sentences from the source. \newcite{pointing} proposed a pointing method to handle the OOV words for summarization and MT. 
In contrast, \textsc{CopyNet} is more general, and not limited to a specific task or OOV words. Moreover, the softmax\textsc{CopyNet} is more flexible than gating in the related work in handling the mixture of two modes, due to its ability to adequately model the content of copied segment. \vspace{-3pt}

%Our work is also inspired by the explicit design on combining content-based addressing together with location-based addressing, which is explored in~\cite{graves2014neural,kurach2015neural}. In addition,  combining ``hard operation" like the copying mechanism in neural networks is also researched for question answering with knowledge bases in~\cite{neelakantan2015neural,yin2015neural2,yin2015neural}
%\vspace{-5pt}
\section{Conclusion and Future Work}
\vspace{-5pt}
%We proposed \textsc{CopyNet} to incorporate copying into the sequence-to-sequence learning framework.  \textsc{CopyNet} can nicely integrate the regular way of word generation in the decoder with the new copying mechanism which can choose sub-sequences in the input sequence and put them at proper places in the output sequence. Our empirical study on both synthetic data sets and real world data sets demonstrates the efficacy of \textsc{CopyNet}. 
We proposed \textsc{CopyNet} to incorporate copying into 
%neural network-based 
the sequence-to-sequence learning  framework.  %Our empirical study on both synthetic data sets and real world data sets demonstrates the efficacy of \textsc{CopyNet}. 
For future work, we will extend this idea to the task where the source and target are in  heterogeneous types, for example, machine translation.

\vspace{-3pt}
\section*{Acknowledgments}
\vspace{-5pt}
This work is supported in part by the China National 973 Project 2014CB340301.

\bibliography{acl2016}
\bibliographystyle{acl2016}

\end{document}